\documentclass[10pt,twocolumn,letterpaper]{article}

\usepackage[final]{cvpr}      

\definecolor{cvprblue}{rgb}{0.21,0.49,0.74}
\usepackage[pagebackref,breaklinks,colorlinks,allcolors=cvprblue]{hyperref}
\usepackage{multirow}
\usepackage{booktabs}  
\usepackage{graphicx}  
\usepackage{float}     

\def\paperID{200} 
\def\confName{CVPR}
\def\confYear{2025}

\title{\textit{ORIDa}: Object-centric Real-world Image Composition Dataset}

\author{
Jinwoo Kim, Sangmin Han, Jinho Jeong, Jiwoo Choi, Dongyeong Kim, Seon Joo Kim\\
Yonsei University\\
{\tt\small \{jinwoo-kim, seonjookim\}@yonsei.ac.kr}\\
\small \url{https://hello-jinwoo.github.io/orida}
}

\begin{document}

\twocolumn[{%
\renewcommand\twocolumn[1][]{#1}%
\maketitle
\vspace{-1cm}
\begin{center}
    \centering
    \includegraphics[width=\textwidth]{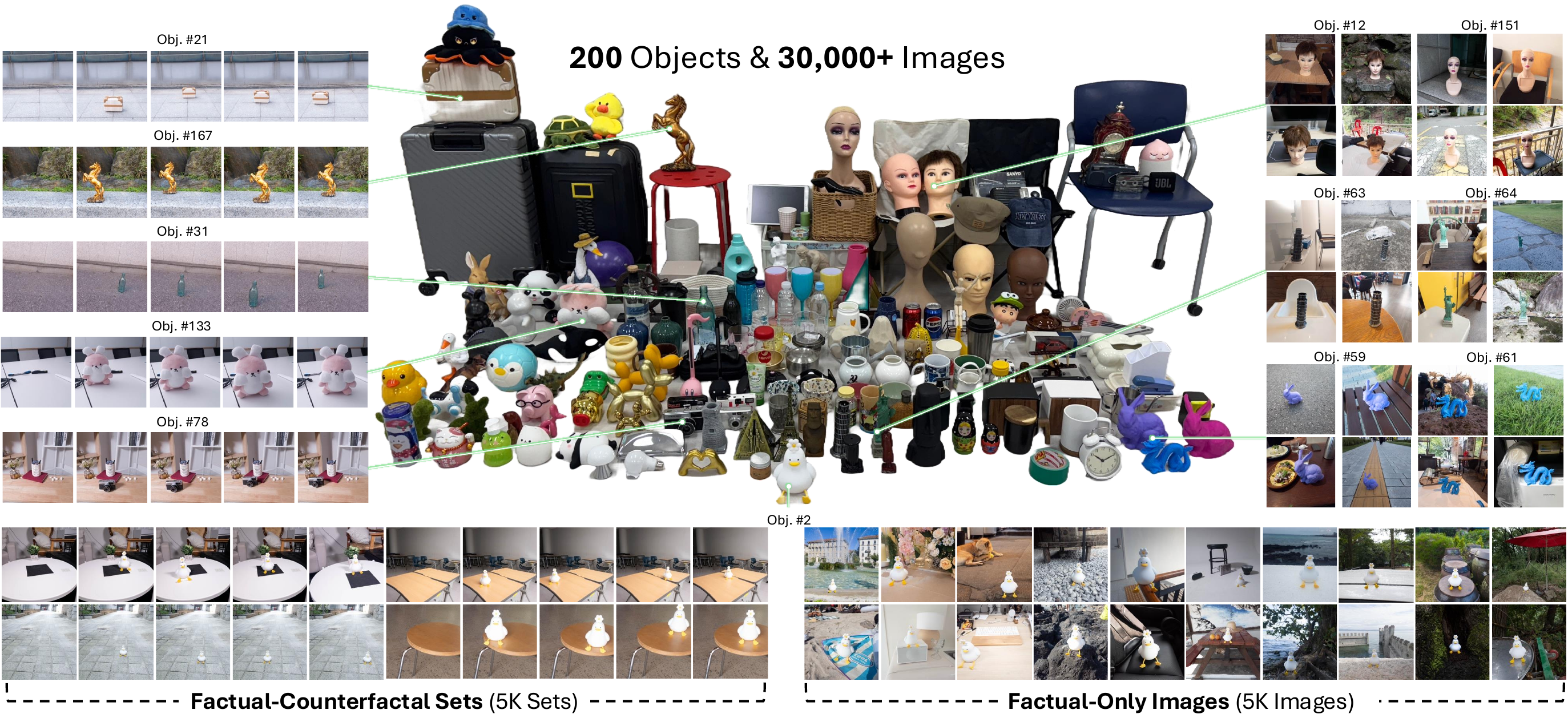}  
    \captionof{figure}{\textbf{Overview of ORIDa.} ORIDa contains 200 unique objects and over 30,000 real-captured images, including factual-counterfactual (F-CF) sets and factual-only (F-Only) images. F-CF sets consist of five images: one background-only and four with the object in different positions. F-Only images capture objects in diverse scenes, enhancing the diversity of the dataset  for object reposition tasks.}
    \label{fig:teaser}
\end{center}

\vspace{-0.3em}
}]
\begin{abstract}
Object compositing, the task of placing and harmonizing objects in images of diverse visual scenes, has become an important task in computer vision with the rise of generative models.
However, existing datasets lack the diversity and scale required to comprehensively explore real-world scenarios. 
We introduce ORIDa (Object-centric Real-world Image Composition Dataset), a large-scale, real-captured dataset containing over 30,000 images featuring 200 unique objects, each of which is presented across varied positions and scenes. 
ORIDa has two types of data: factual-counterfactual sets and factual-only scenes. 
The factual-counterfactual sets consist of four factual images showing an object in different positions within a scene and a single counterfactual (or background) image of the scene without the object, resulting in five images per scene.
The factual-only scenes include a single image containing an object in a specific context, expanding the variety of environments. 
To our knowledge, ORIDa is the first publicly available dataset with its scale and complexity for real-world image composition.
Extensive analysis and experiments highlight the value of ORIDa as a resource for advancing further research in object compositing.  

\end{abstract}    
\section{Introduction}
\label{sec:intro}

\begin{table*}[htbp]
    \centering
    \renewcommand{\arraystretch}{1.3}  
    \caption{
        \textbf{Comparison of datasets for compositional image editing tasks.}
        obj. and b.g. stand for object and background, respectively.
    }
    \vspace{0.2em}
    \resizebox{\textwidth}{!}{%
    \begin{tabular}{c|cccccccccccc}
        \toprule
        \multirow{2}{*}{Dataset} & \multirow{2}{*}{Target task} & \multirow{2}{*}{\# of data} & \multirow{2}{*}{\begin{tabular}[c]{@{}c@{}}\# of\\ objects\end{tabular}} & \multirow{2}{*}{\begin{tabular}[c]{@{}c@{}}\# of scenes\\ per an obj.\end{tabular}} & \multirow{2}{*}{\begin{tabular}[c]{@{}c@{}}\# of pos.\\ per a scene\end{tabular}} & \multirow{2}{*}{Real} & \multirow{2}{*}{\begin{tabular}[c]{@{}c@{}}Factual\\ (including obj.)\end{tabular}} & \multirow{2}{*}{\begin{tabular}[c]{@{}c@{}}Counterfact.\\ (b.g. only)\end{tabular}} & \multirow{2}{*}{\begin{tabular}[c]{@{}c@{}}Train\\ set\end{tabular}} & \multirow{2}{*}{\begin{tabular}[c]{@{}c@{}}Test\\ set\end{tabular}} & \multirow{2}{*}{Public} & \multirow{2}{*}{DNG} \\
         &  &  &  &  &  &  &  &  &  &  &  &  \\ \hline \hline
        COCOEE \cite{yang2023paint} & \begin{tabular}[c]{@{}c@{}}Examplar-based\\ Image Editing\end{tabular} & 2.5K triplets & 2.5K & 1 & 1 & O & O & X & X & O & O & X \\ \hline
        DreamEdit \cite{li2023dreamedit} & \begin{tabular}[c]{@{}c@{}}Subject-driven\\ Image Editing\end{tabular} & \begin{tabular}[c]{@{}c@{}}440 images\\ (source-only)\end{tabular} & 22 & 20 & 1 & - & X & O & X & O & O & X \\ \hline
        DreamBooth \cite{ruiz2023dreambooth} & \begin{tabular}[c]{@{}c@{}}Subject-driven\\ Image Generation\end{tabular} & 157 images & 30 & 4$\sim$6 & 1 & O & O & X & X & O & O & X \\ \hline
        FOS-Com \cite{zhang2023controlcom} & \begin{tabular}[c]{@{}c@{}}Object\\ Compositing\end{tabular} & 640 triplets & 640 & 1 & 1 & X & O & X & X & O & O & X \\ \hline
        ObjectDrop \cite{winter2024objectdrop} & \begin{tabular}[c]{@{}c@{}}Object\\ Compositing\end{tabular} & \begin{tabular}[c]{@{}c@{}}5K images\\ (2.5K pairs)\end{tabular} & 2.5K & 1 & 1 & O & O & O & O & X & X & X \\ \hline
        \textbf{\begin{tabular}[c]{@{}l@{}}ORIDa (Ours)\end{tabular}} & \begin{tabular}[c]{@{}c@{}}Object\\ Compositing\end{tabular} & \textbf{\begin{tabular}[c]{@{}c@{}}30K images\\ (5K sets + 5K images)\end{tabular}} & \textbf{200} & \textbf{$\approx$50} & \textbf{1$\sim$4} & \textbf{O} & \textbf{O} & \textbf{O} & \textbf{O} & \textbf{O} & \textbf{O} & \textbf{O} \\ 
        \bottomrule
    \end{tabular}%
    }
    \label{tab:datasets_comparison}
\end{table*}

\vspace{-0.5em}

Object compositing, or image composition, refers to the task of placing objects into visual scenes in a manner that preserves realism and contextual consistency. 
This task is critical in many computer vision applications, including image editing, augmented reality, and scene understanding, where objects must seamlessly blend into complex environments. 
The difficulty lies in ensuring that objects not only fit naturally into a wide range of scenes but also retain their identity and appearance. 
Successfully addressing object compositing involves overcoming key challenges, such as maintaining the object’s identity, harmonizing its appearance with the scene, and managing complex factors like lighting, shadows, and geometric alignment. 

Recent advancements in object compositing can be broadly categorized into two groups: training-free methods and training-based approaches. 
Training-free methods \cite{chen2024freecompose,lu2023tf} have delivered impressive results, generating object placements without the need for task-specific datasets. 
Despite their success, these methods often struggle with fine details, such as scene harmonization and preserving object identity. 
Training-based approaches \cite{song2023objectstitch,winter2024objectdrop}, in contrast, benefit from data-driven training and can be further divided into those using synthetic data and those using real-world data. 
While training with synthetic data \cite{song2023objectstitch,yang2023paint} has significantly advanced object compositing, the lack of real-world complexity limits the realism of the generated images. 
ObjectDrop \cite{winter2024objectdrop} has successfully enhanced object compositing using real-captured data; however, its limited scale and scene variability per object necessitate the incorporation of large-scale synthetic datasets. 
Additionally, the ObjectDrop dataset is not publicly available.

To this end, we present \textbf{\textit{ORIDa}} (\textbf{O}bject-centric \textbf{R}eal-world \textbf{I}mage Composition \textbf{Da}taset), the first large-scale real-captured public dataset specifically designed for the object compositing. 
ORIDa contains over \textbf{30,000} images of \textbf{200} unique objects, each placed in an average of \textbf{50} diverse scenes, providing an extensive and varied dataset for studying object placement in real-world contexts. 
ORIDa offers both \textit{factual-counterfactual} sets \cite{lewis2013counterfactuals,winter2024objectdrop} in which each object is captured in four different positions per scene, alongside a corresponding scene without the object, and \textit{factual-only} images, enriching the dataset with a wide range of contextual possibilities. 
To ensure the dataset quality, particular attention was given during data collection process in order to minimize external factors beyond the object’s presence. 

Compared to existing datasets, ORIDA offers several key advantages as shown in \Cref{tab:datasets_comparison}.
Datasets like COCOEE \cite{yang2023paint} and FOS-Com \cite{zhang2023controlcom} are designed solely for benchmarking and consist of object compositing sets derived from existing datasets, not captured specifically for the task. 
While DreamEditBench \cite{li2023dreamedit} and DreamBooth Dataset \cite{ruiz2023dreambooth} offer more variation with multiple image composition scenarios per object, they are also intended only for testing, making it less suited for advancing models. 
ObjectDrop \cite{winter2024objectdrop} consists of real-captured data suitable for training, however, it includes only one image pair per object and is not publicly available. 
In contrast, ORIDA is a large-scale, real-captured and publicly available dataset that offers multiple images per object across varied scenes, making it a more versatile resource for both training and evaluation.

We provide a detailed analysis on ORIDa, demonstrating the wide range of object categories, visual attributes, and contextual variations captured in the dataset. 
In addition, we show experimental results on object removal and object insertion tasks using a fine-tuned model, trained on ORIDa without incorporating any synthetic datasets.
The results validate the potential of our dataset to support realistic object compositing in diverse scenarios, enabling future exploration in compositional image generation and editing.
\begin{figure*}[htbp]
    \centering
    \includegraphics[width=\textwidth]{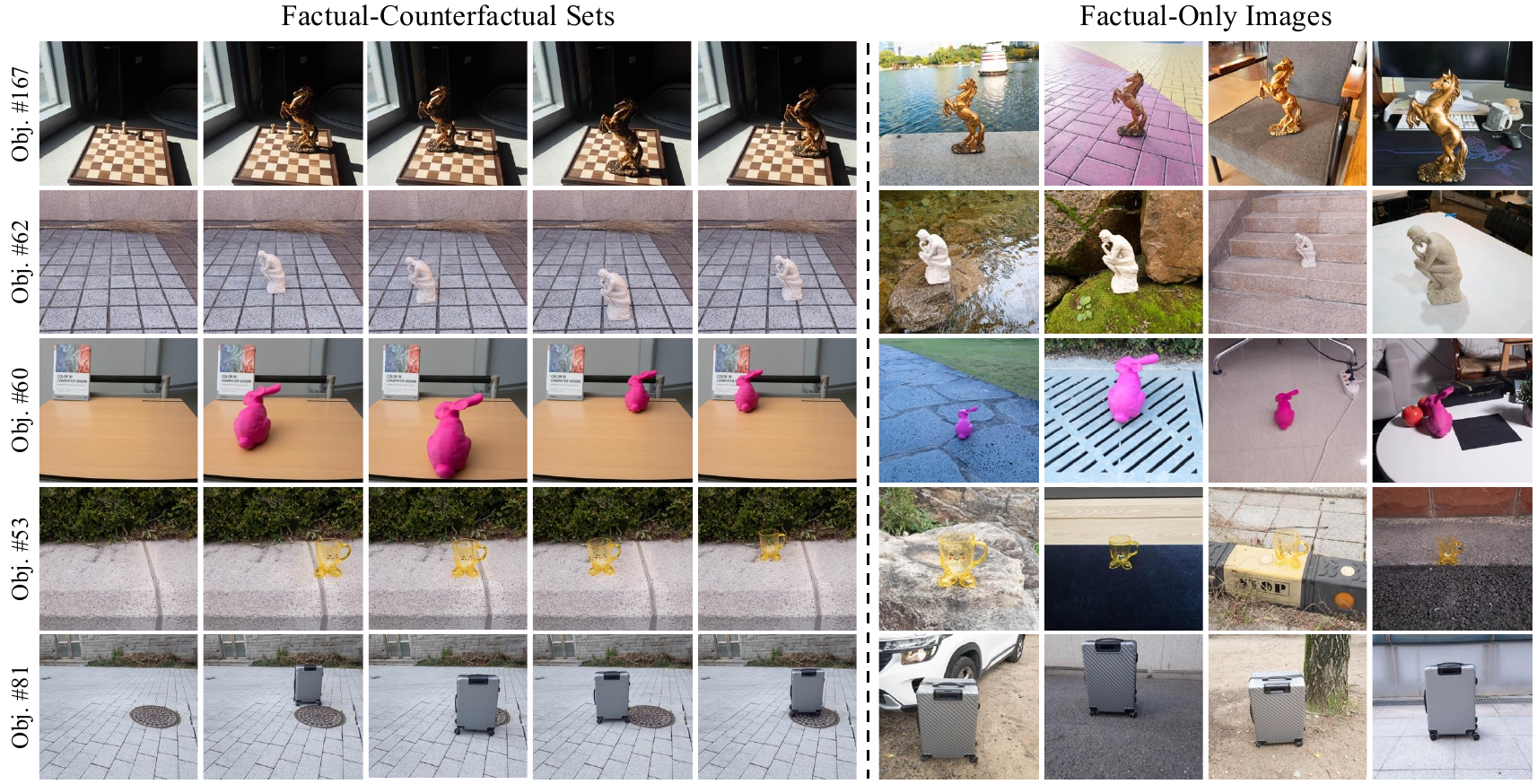}
    \caption{\textbf{Examples of Factual-Counterfactual (F-CF) Sets and Factual-Only (F-Only) Images.} The left side shows F-CF sets, consisting of one background-only image and four object-inserted images captured with the object in different positions. The right side displays F-Only images, which feature objects in diverse scenes without corresponding background-only images.}
    \label{fig:data_samples}
\end{figure*}


\section{Related Work}
\label{sec:related_works}

\subsection{Object Compositing}
\label{subsec:object_compositing}

Key challenges in object compositing include identity preservation \cite{song2024imprint,ruiz2023dreambooth,wu2024infinite,cai2021frequency}, color harmonization \cite{guerreiro2023pct,jiang2021ssh,ke2022harmonizer,xue2022dccf,cong2021bargainnet,cong2020dovenet}, shadow generation/removal \cite{hong2022shadow,liu2020arshadowgan,liu2021shadow,wang2018stacked,le2020shadow,hu2019mask}, and geometric correction \cite{yenphraphai2024image,wang2024diffusion,sajnani2024geodiffuser,kwon2024geometry}.
In contrast to earlier approaches that tend to address these challenges individually, recent methods \cite{song2023objectstitch,winter2024objectdrop,zhang2023controlcom,lu2023dreamcom,li2023dreamedit,sarukkai2024collage,chen2024anydoor,yang2023paint,lu2023tf} aim to handle them within unified frameworks with the advent of diffusion models \cite{songdenoising,songscore,ho2020denoising,rombach2022high,podellsdxl}. 
For example, models like ObjectStitch \cite{song2023objectstitch} and Paint-by-Example \cite{yang2023paint} attempt to integrate geometric alignment, harmonization, and identity preservation, although issues such as object fidelity and accurate shadow modeling remain.

Existing approaches to object compositing can be classified into training-free and training-based methods. 
Training-free methods \cite{lu2023tf,chen2024freecompose} leverage pre-trained models and do not require task-specific datasets.
For example, FreeCompose \cite{chen2024freecompose} employs a mask-guided loss function during inference, which harmonizes the inserted object with the given background scene.
However, they often struggle in maintaining realism and preserving object identity in complex environments. 
On the other hand, training-based methods \cite{song2024imprint,song2023objectstitch,winter2024objectdrop,zhang2023controlcom,yang2023paint}, which depend on large datasets, have demonstrated considerable potential for enhancing object compositing performance. 
Due to the lack of real-world image composition datasets, many of these methods resort to generating synthetic data. 
This process involves masking an object’s area and refilling the masked part with an augmented version of the target object to train their models. 
While synthetic data helps to overcome data scarcity, it often lacks the complexity and diversity of real-world scenes, which can restrict the models’ performance.

\subsection{Datasets for Image Composition}
\label{subsec:image_compositing_datasets}

Various datasets support the object compositing task. 
COCOEE \cite{yang2023paint} and FOS-Com \cite{zhang2023controlcom} are designed as benchmarking datasets.
While useful for testing models, these datasets are not specifically captured for object compositing, which limits their effectiveness for model training and exploration of object placement variations. 
Furthermore, they offer only one compositing set per object.

DreamEditBench \cite{li2023dreamedit} provides around 20 compositing sets per object. 
Nevertheless, like COCOEE and FOS-Com, it is designed solely for benchmarking and includes only source objects and background images, lacking ground truth object-included images. 
Similarly, DreamBooth Dataset \cite{ruiz2023dreambooth} includes 30 subjects, comprising both objects and live subjects/pets. 
Despite offering some variation, DreamBooth Dataset is also intended solely for evaluation, and its relatively small scale, along with the absence of counterfactual (or background) images, limits its usefulness for training object compositing models.

The dataset most similar to ORIDa is ObjectDrop \cite{winter2024objectdrop} which consists of 2,500 real-captured factual-counterfactual pairs, enabling the training of object compositing models. 
However, ObjectDrop also relies on synthetic datasets to train object insertion models.
Furthermore, since the dataset is not publicly available, it is impossible to use it for developing broader models.
\begin{figure}[t]
    \centering
    \includegraphics[width=\linewidth]{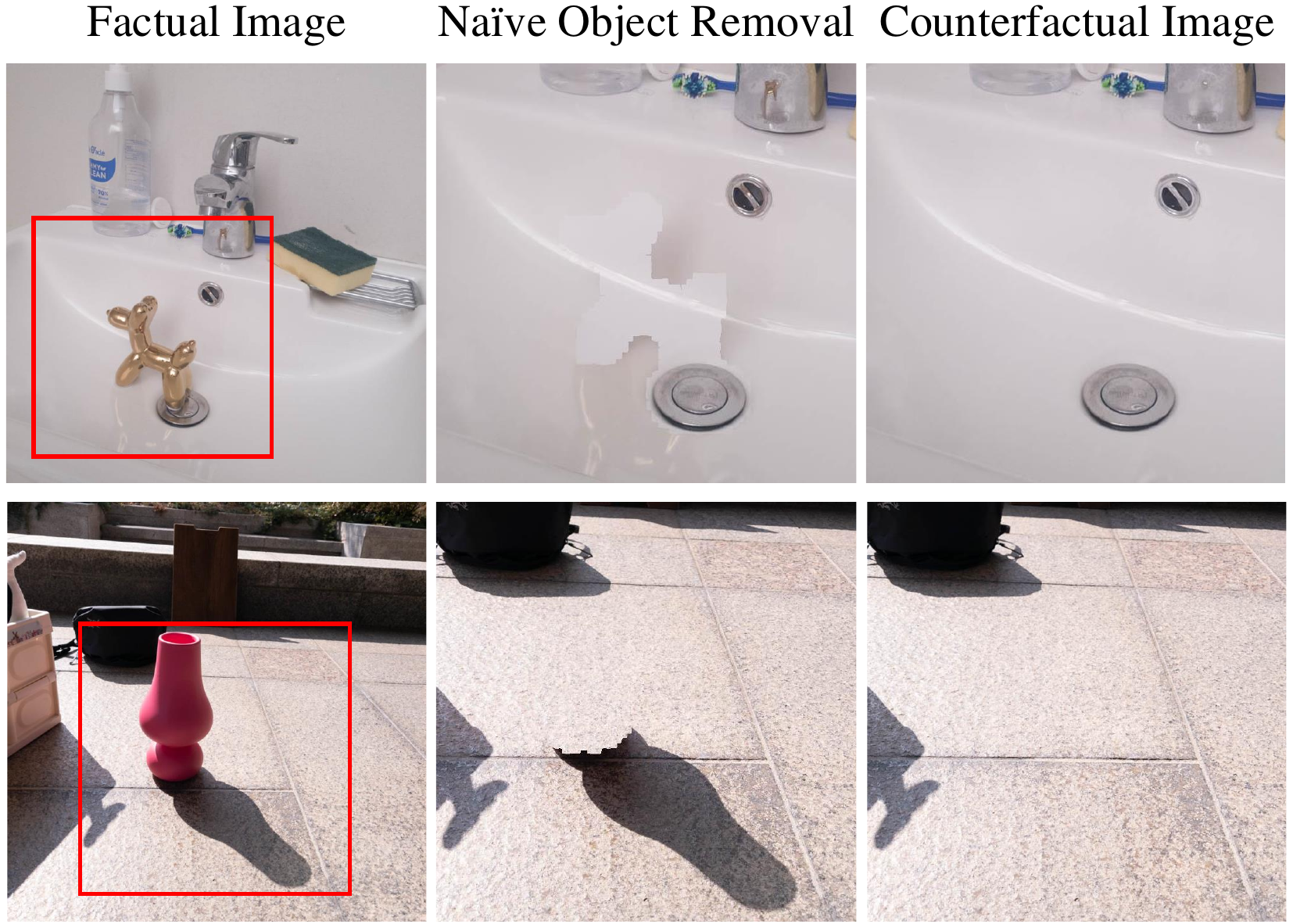}
    \vspace{-1.8em}
    \caption{\textbf{Visualization of factual-counterfactual concept.} The first column shows factual images with objects present in the scene. The second column displays naïve object removal results, where background is synthetically stitched into the object region of the factual image. The third column presents the corresponding ground-truth counterfactual images, which consist solely of the background without the object. These examples demonstrate the object-to-scene effects, including shadows and reflections.}
    \label{fig:factual_counterfactual}
\end{figure}

\section{Dataset Collection}
\label{sec:dataset_collection}

\subsection{Objects}
\label{subsec:objects_introduction}
Our dataset includes a total of 200 unique objects.
To ensure consistency and maintain the focus on the object’s placement within and across scenes, we limit the variation in object poses during data capture. 
This allows for more controlled analysis and model training, centering on the object’s interaction with its environment rather than pose dynamics.
More detailed information about the objects and their characteristics will be elaborated in \Cref{sec:dataset_statistics}.

\subsection{Image Capture}
\label{subsec:data_capture}

We used five different cameras to capture images: Galaxy S10, Galaxy S20, Galaxy S22, Galaxy S24, and Galaxy Note10+, all in PRO mode to obtain raw DNG files. 
Samples are provided in \Cref{fig:data_samples}.

\begin{figure}[t]
    \centering
    \includegraphics[width=\linewidth]{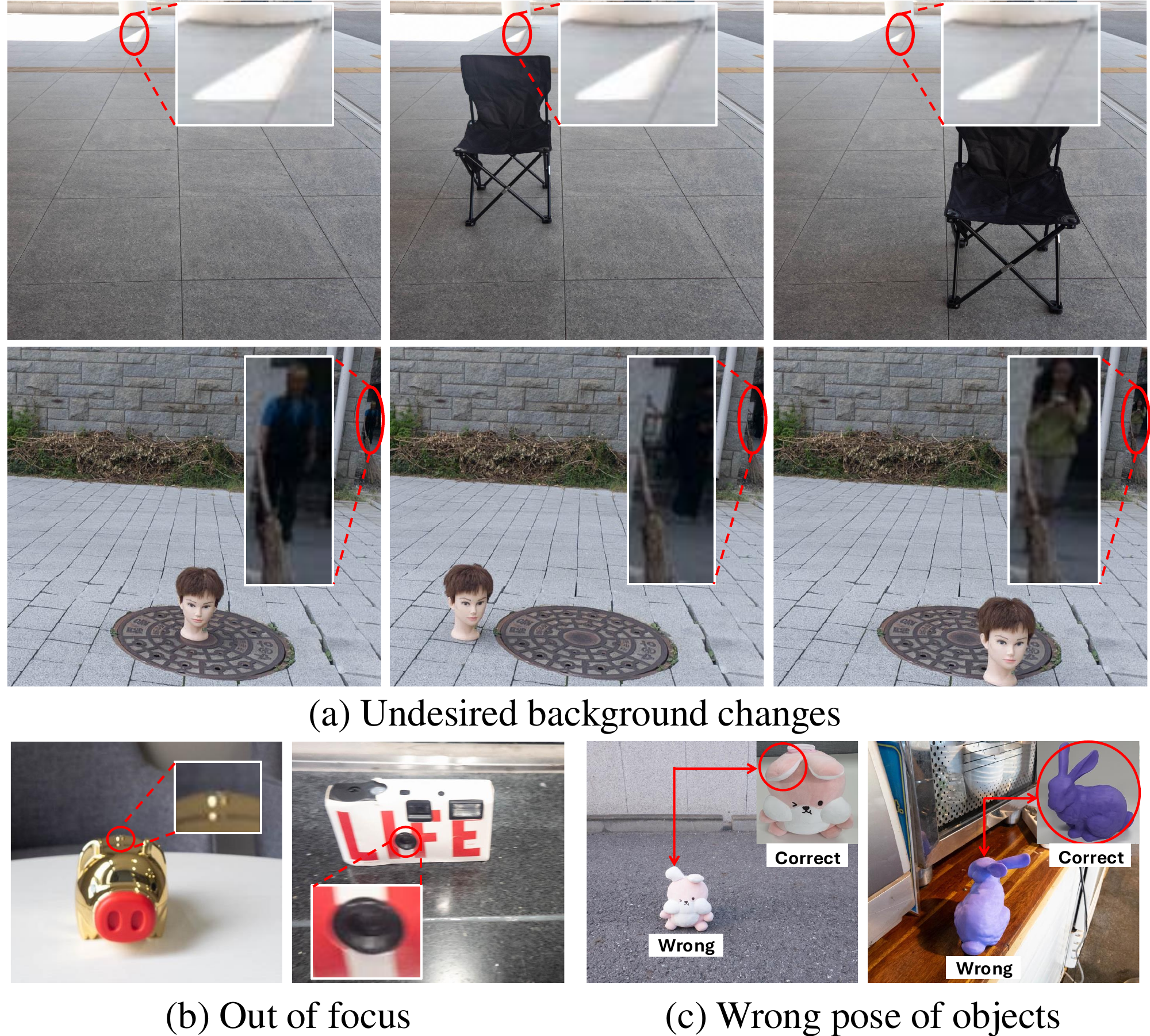}
    \vspace{-1.7em}
    \caption{\textbf{Filtering criteria.} Common issues include: (a) undesired background changes such as lighting shifts or pedestrians, (b) out-of-focus images, and (c) incorrect object poses.} 
    \label{fig:filtering}
\end{figure}
\begin{figure}[t]
    \centering
    \includegraphics[width=\linewidth]{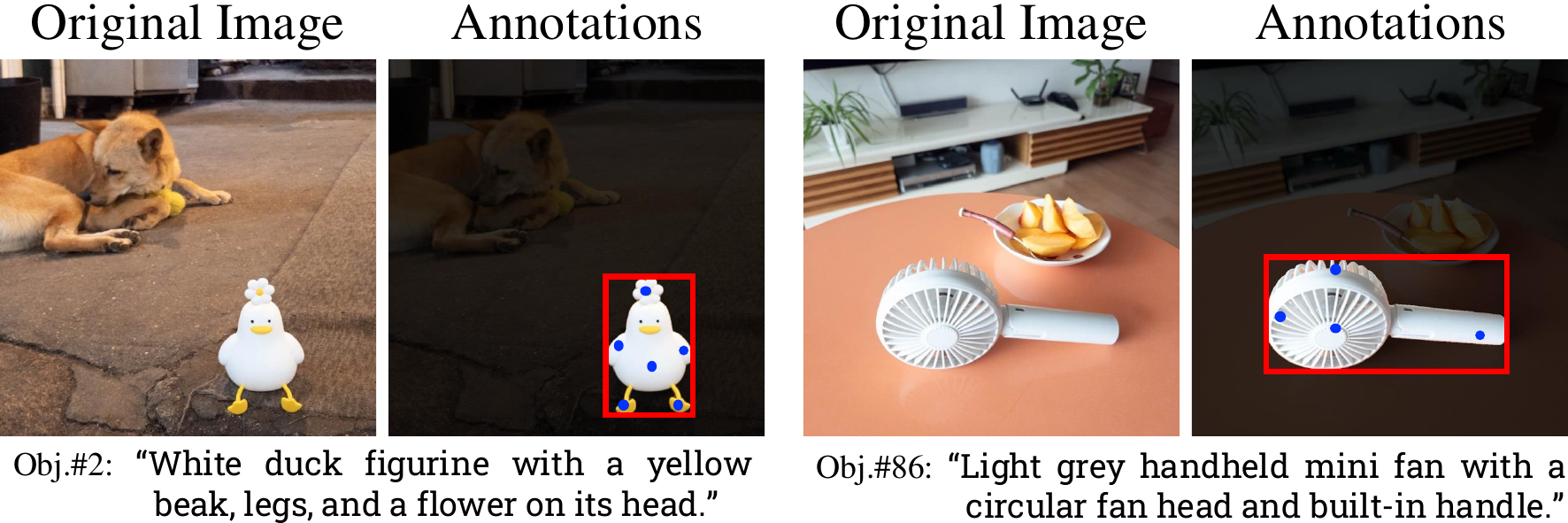}
    \vspace{-1.5em}
    \caption{\textbf{Annotation examples.} Each object includes detailed annotations such as captions, object points, bounding boxes, and segmentation masks.}
    \label{fig:annotations}
\end{figure}
\begin{figure*}[htbp]
    \centering
    \includegraphics[width=\textwidth]{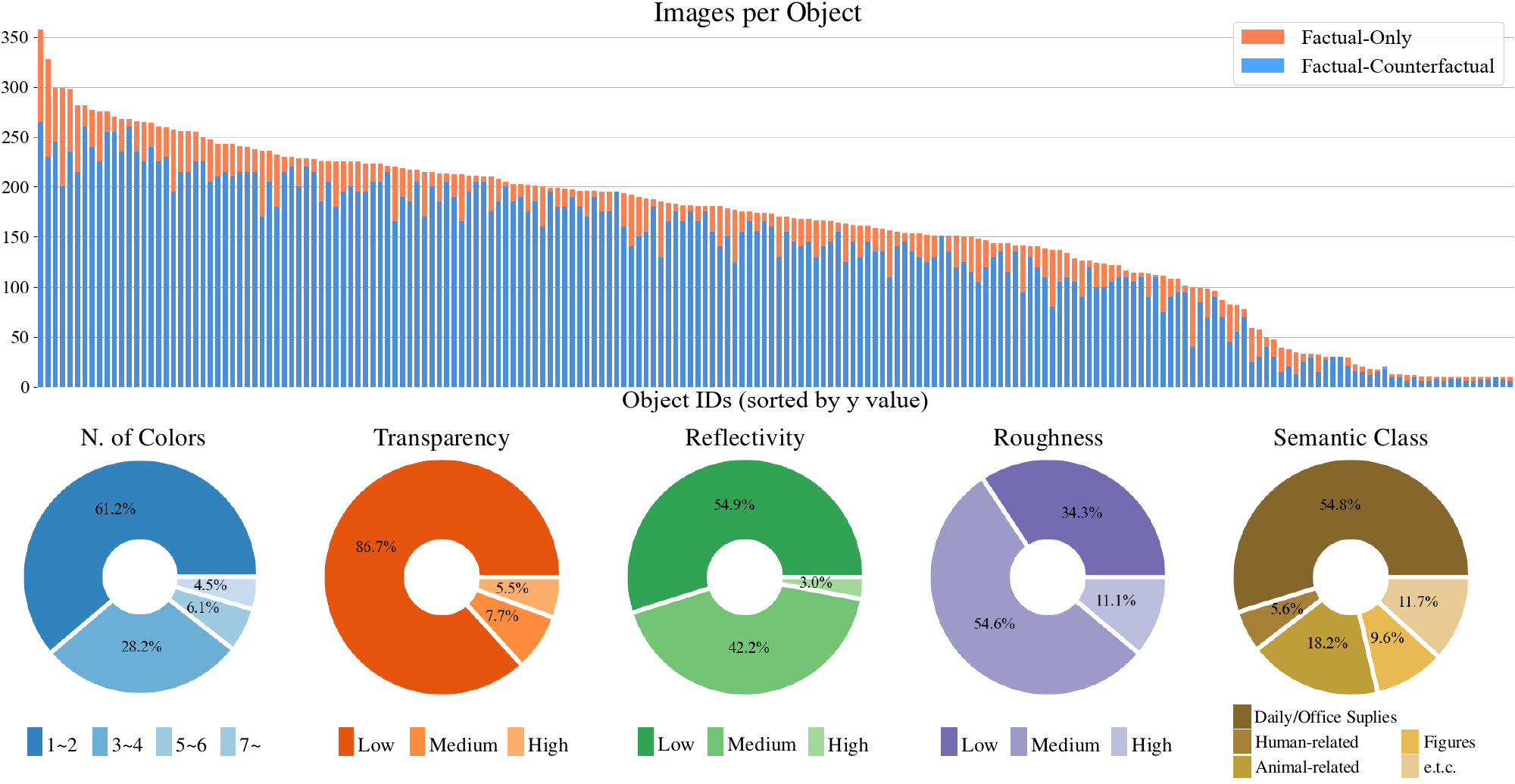}
    \caption{\textbf{Dataset statistics per object and attribute.} The top chart displays the number of images per object, sorted by y-value, for both factual-only and factual-counterfactual sets. The bottom charts present the percentage distribution of objects based on key attributes: number of colors, transparency, reflectivity, roughness, and semantic class, illustrating the variety and diversity within the dataset.}
    \label{fig:n_obj}
\end{figure*}

\noindent \textbf{Factual-Counterfactual (F-CF) Sets.}
The concept of F-CF sets is inspired by ObjectDrop \cite{winter2024objectdrop}: ``if the object did not exist, this reflection would not occur" (\Cref{fig:factual_counterfactual}).
However, our dataset differs in two key aspects: (1) our dataset includes multiple scenes for each object, whereas ObjectDrop provides only a single scene, and (2) covers multiple positions of an object within a given background, while ObjectDrop offers only one object position.
As a result, each F-CF set in ORIDa consists of five images: one background-only image and four images with the object in different positions while fixing the background. 
To collect F-CF sets, we carefully selected shooting locations by considering consistent lighting conditions, stable backgrounds, and diverse scenes.
To ensure consistency within each set, we fixed key camera settings such as shutter speed, ISO, WB, and focus, during a single capture process, preserving the natural lighting and scene conditions across images in each set.
Cameras were fixed on tripods and we captured a series of five consecutive images with remote controllers to maintain stability of the camera position.

\noindent \textbf{Factual-Only (F-Only) Images.}
We also collected F-Only images to increase scene diversity. 
These images are comparatively easier to capture since they do not require separate background shots, allowing us to efficiently gather object-included images across a variety of backgrounds.

\subsection{Data Filtering}
\label{subsec:data_filtering}
To ensure that any variations in the scenes are solely due to the presence of objects and to uphold the overall image quality, we meticulously inspected all captured images. 
We identified several undesired cases, as illustrated in \Cref{fig:filtering}: (1) unintended background changes, (2) out-of-focus images, and (3) incorrect object poses.
In addition to these cases, we performed inspections to identify and filter out other inconsistencies or defects. 
We selected 5,699 F-CF sets from the initial 7,000 F-CF sets and retained 5,035 F-Only images from the original 5,500 F-Only images.

\subsection{Annotations}
\label{subsec:annotations}
In addition to the filtered images, we provide comprehensive annotations for the dataset, including captions for 200 individual objects, object points, bounding boxes, and segmentation masks as shown in \Cref{fig:annotations}. 

For generating object captions, we captured object-centric images with simple backgrounds where the objects are dominant. 
These images are then used as inputs for GPT-4o \cite{openai2024gpt4o} and Gemini 1.5 Pro \cite{google2024gemini} to create object descriptions.
For localization-related annotations, such as bounding boxes and segmentation masks, we manually annotated points for each target object in the images. 
These annotated points are subsequently fed into SAM2 \cite{ravi2024sam} to generate precise segmentation masks and bounding boxes. 
All images in our dataset, excluding background-only images, include localization-related annotations.

Moreover,raw DNG files in ORIDa provide flexibility for additional ISP (Image Signal Processing) augmentations. 
This feature is crucial for effective harmonization in object compositing, allowing for exploration with different color and lighting conditions on the original raw files.
\begin{figure}[t]
    \centering
    \includegraphics[width=\linewidth]{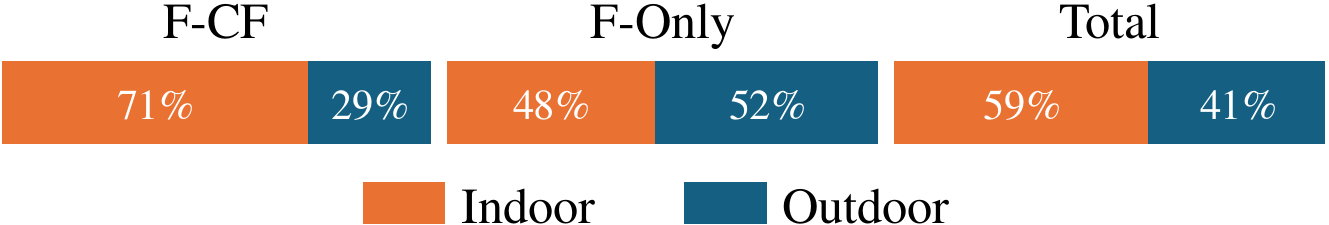}
    \vspace{+0.02em}
    \caption{\textbf{Indoor/Outdoor ratio.} Distribution of indoor and outdoor scenes for factual-counterfactual sets (F-CF), factual-only images (F-Only), and the entire dataset. }
    \label{fig:indoor_outdoor}
\end{figure}

\begin{figure*}[htbp]
    \centering
    \includegraphics[width=\textwidth]{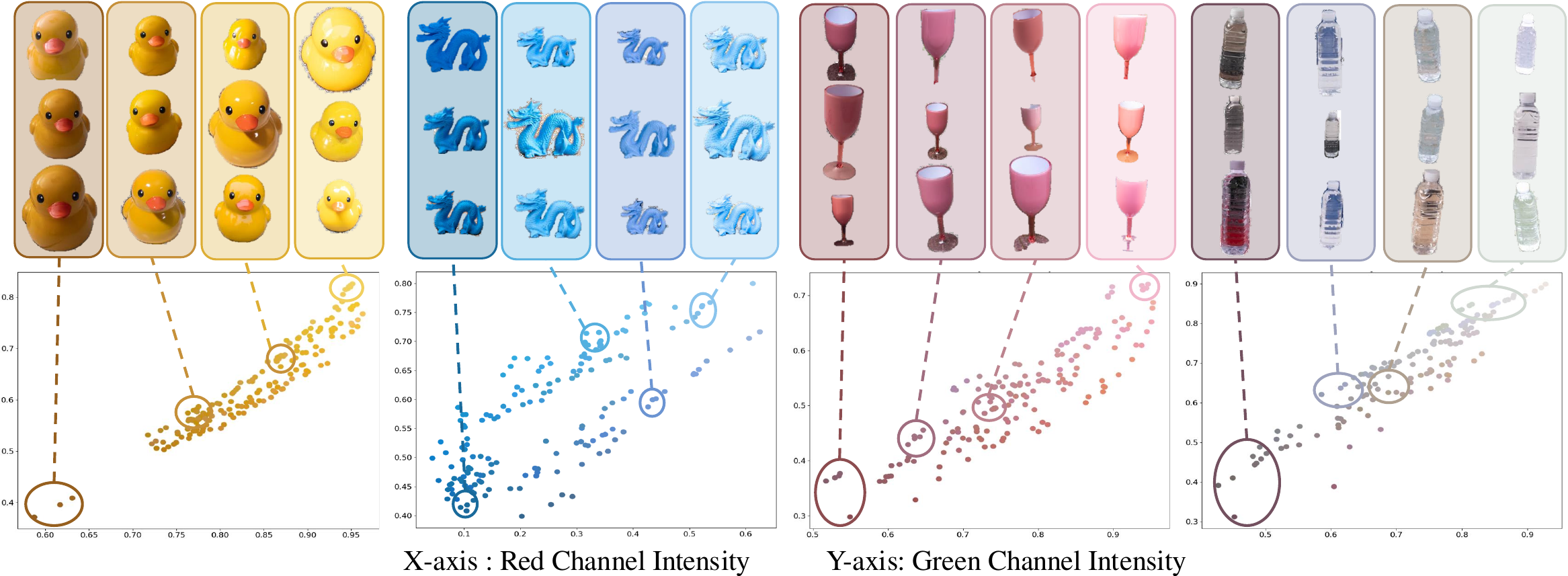}
    \vspace{-1.7em}
    \caption{\textbf{Visualization of color variations across objects.} Example objects are shown with their respective color distributions plotted based on red and green channel intensities. Each plot highlights how the appearance of objects varies under different lighting conditions and backgrounds, illustrating the dataset’s ability to capture diverse visual contexts.}
    \label{fig:object_color}
\end{figure*}

\section{Dataset Statistics}
\label{sec:dataset_statistics}

We present several statistics and analyses of our dataset. 
\Cref{fig:n_obj} (top) illustrates the distribution of image counts per object both for the F-CF and F-Only.
Note that the y-axis represents the number of images, with F-CF counts calculated by multiplying the number of sets by five.
Additionally, as shown in \Cref{fig:n_obj} (bottom), we categorize objects based on five attributes: number of colors, transparency, reflectivity, roughness, and semantic classes. 
These attributes help to understand the variety of visual properties and textures in the dataset. 
The majority of objects have between one and four main colors, while attributes like transparency and reflectivity are distributed across low to medium levels, indicating a range of visual complexity.
Objects are also classified into various semantic classes, such as daily/office supplies, human-related items, and animal-related objects.

Furthermore, our dataset includes scenes captured both indoors and outdoors, providing a mix of environments for object compositing. 
We calculated the ratio of indoor to outdoor scenes separately for F-CF sets, F-Only images, and the entire dataset. 
As shown in \Cref{fig:indoor_outdoor}, the dataset maintains a balanced distribution between indoor and outdoor settings, with 41\% of images captured outdoors. 

A unique characteristic of ORIDa is that it captures both the object-to-scene effects and scene-to-object effects. The former refers to how an object impacts its environment such as shadows and reflections as shown in \Cref{fig:factual_counterfactual}, while the latter considers how varying contexts affect an object’s appearance.
To explore the diversity in object appearances, we analyze the mean color values of some objects across varied scenes. 
\Cref{fig:object_color} plots the mean color distributions, showing how the appearance of objects shifts under varying lighting conditions and backgrounds. 
The plots demonstrate that while objects generally maintain their defining characteristics, there are noticeable changes in color intensities based on the context, illustrating the dataset’s ability to represent a wide range of appearances for the same object.


\section{Experiments}
\label{sec:experimnets}

\begin{figure*}[htbp]
    \centering
    \includegraphics[width=0.96\textwidth]{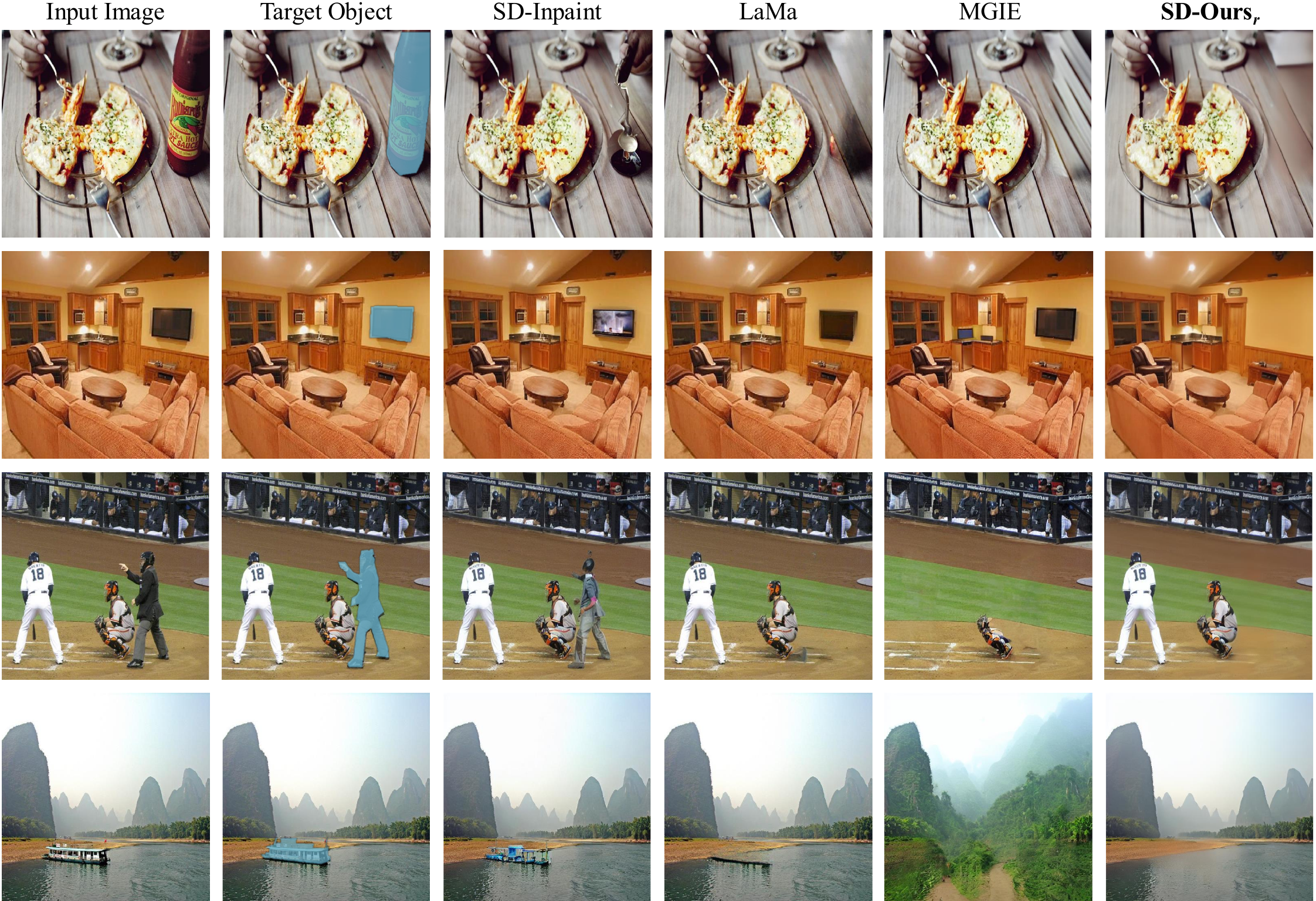}
    \vspace{-0.6em}
    \caption{\textbf{Object removal - qualitative results across different methods.} SD-Inpaint \cite{rombach2022high}, LaMa \cite{suvorov2022resolution}, MGIE \cite{fu2023guiding}, and SD-Ours. For MGIE, a text prompt such as “remove the hot sauce from the photo” (for the first row) is used to instruct the model.}
    
    \label{fig:removal}
\end{figure*}

\subsection{Experimental Settings}
\label{sec:experimental_settings}

\noindent \textbf{Train datasets.}
To enhance the diversity of the training data, we utilize the raw files in ORIDa and apply ISP augmentations using Adobe Lightroom. 
Five different ISP settings are applied: (1) as-shot, (2) higher temperature, (3) lower temperature, (4) higher vibrance, and (5) lower vibrance. 
For the object insertion task, we use additional 60,000 images from the COCO dataset \cite{lin2014microsoft}, paired with 250,000 object masks, to train the model to maintain the identity of source objects. 
Please note that we only utilize the original images from the COCO dataset without any synthetic data, freeing us from the hyper-parameters and complex recipes required for data synthesis \cite{song2023objectstitch,winter2024objectdrop,yang2023paint,zhang2023controlcom}.

\noindent \textbf{Model.}
We fine-tuned a public pretrained StableDiffusion (SD)-Inpaint \cite{rombach2022high,huggingface2024diffusersinpainting}, for both object removal and insertion task without major modification of its architecture. 
The U-Net \cite{ronneberger2015u} in SD-Inpaint receives a 9-channel input: four channels for the input latent, four channels for the condition latent, and one channel for the target object mask. 

\subsection{Object Removal}
\label{sec:object_removal}

We compare our model (SD-Ours$_r$) with SD-Inpaint \cite{rombach2022high,huggingface2024diffusersinpainting}, LaMa \cite{suvorov2022resolution}, and MGIE \cite{fu2023guiding}. 
We use images from COCO dataset for qualitative results (\Cref{fig:removal}) and user studies (\Cref{tab:removal_user_study}), while an out-held test set from ORIDa is used to evaluate automatic quantitative metrics (\Cref{tab:removal}).

\begin{table}
\centering
\caption{\textbf{Object removal - user studies.} Participants rated object removal results on a scale of 1 to 5 across five criteria: context preservation, effectiveness of object removal, elimination of object-related effects (e.g., shadows, reflections), minimization of artifacts, and overall image quality.}
\resizebox{\linewidth}{!}{%
\begin{tabular}{lcccc}
\toprule
                                & SD-Inpaint & LaMa & MGIE & SD-Ours$_r$ \\ \hline
Rating (Max: 5) $\uparrow$   & 2.78     & 2.63 & 1.96 & 4.23 \\
\bottomrule
\end{tabular}
}%
\label{tab:removal_user_study}
\end{table}
\begin{table}
\centering
\caption{\textbf{Object removal – automatic metrics.} Comparison with the inpainting baseline (SD-Inpaint) on ORIDa held-out test set.}
\vspace{-0.7em}
\resizebox{\linewidth}{!}{%
\begin{tabular}{lcccc}
\toprule
           & PSNR $\uparrow$ & DINO $\uparrow$ & CLIP $\uparrow$ & LPIPS $\downarrow$ \\ \hline
SD-Inpaint   & 21.76 & 0.845 & 0.903 & 0.108 \\
SD-Ours$_r$                    & \textbf{25.60} & \textbf{0.902} & \textbf{0.938} & \textbf{0.088} \\
\bottomrule
\end{tabular}
}%
\label{tab:removal}
\end{table}

As shown in \Cref{fig:removal}, our approach demonstrates better object removal performance.
While SD-Inpaint and LaMa perform reasonably well, they often struggle with erasing shadows and reflections. 
MGIE, which uses text prompts, offers flexibility but can introduce artifacts.
In contrast, SD-Ours$_r$ effectively preserves the visual context by accurately erasing shadows, lighting, and object itself. 
This improved performance can be attributed to training exclusively on ORIDa, which provides diverse and high-quality real-world data to handle complex visual scenarios.
Please note that some softness observed in both the inpainted regions and surrounding background likely results from the limitations inherent in the pretrained model.

In addition, we also provide quantitative results using both user study ratings from 76 randomly selected participants (\Cref{tab:removal_user_study}) and automatic metrics (\Cref{tab:removal}).
These evaluations further validate the effectiveness of our dataset in training a model to achieve realistic object removal.

\begin{figure*}[htbp]
    \centering
    \includegraphics[width=\textwidth]{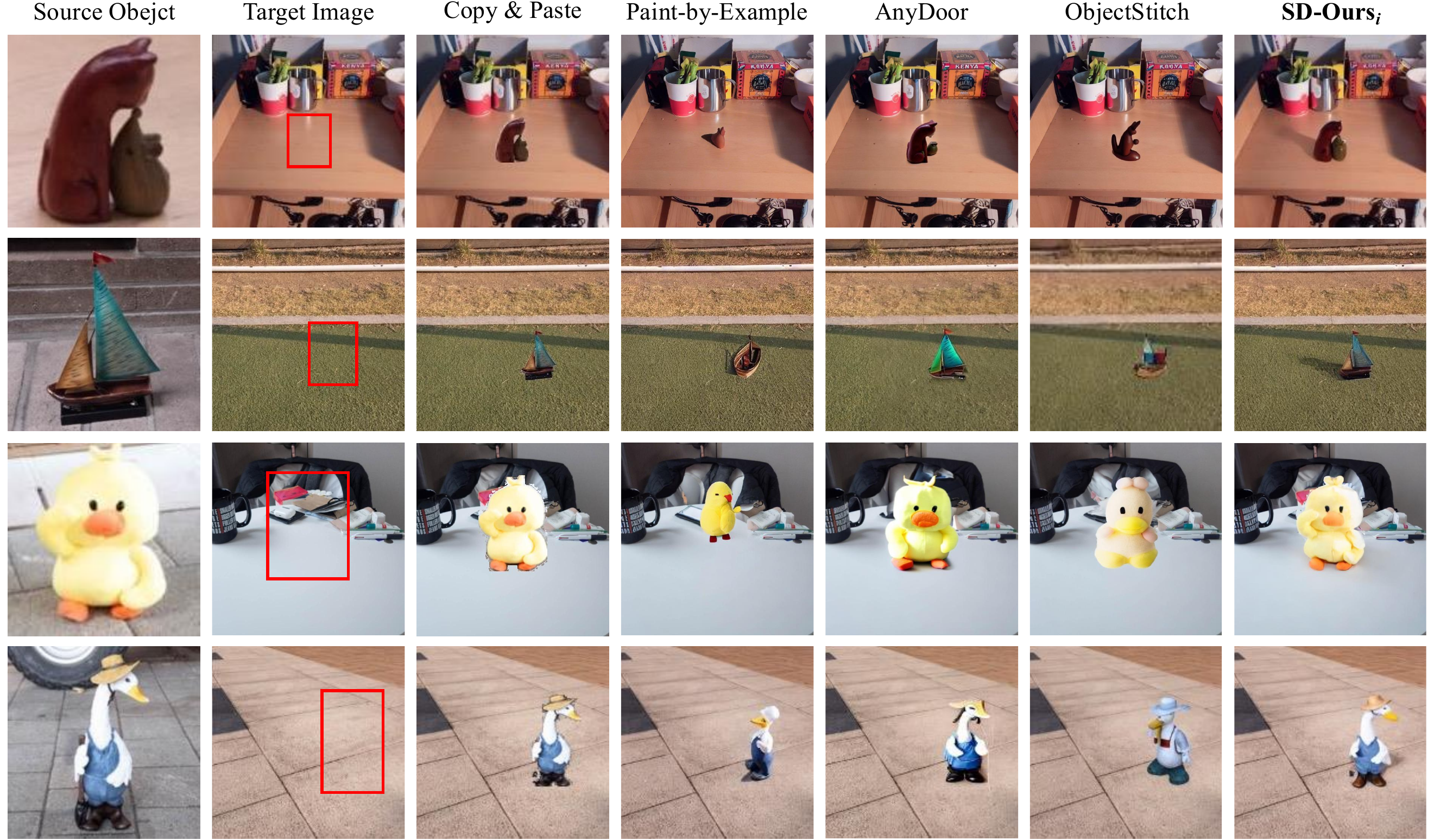}
    \caption{\textbf{Object insertion - qualitative results.} For each row, the Source Object is inserted into the Target Image. Results illustrate differences in identity preservation, shadow generation, color harmonization, and overall realism.} 
    \label{fig:insertion}
\end{figure*}
\subsection{Object Insertion}
\label{sec:object_insertion}

\begin{figure}[t]
    \centering
    \includegraphics[width=\linewidth]{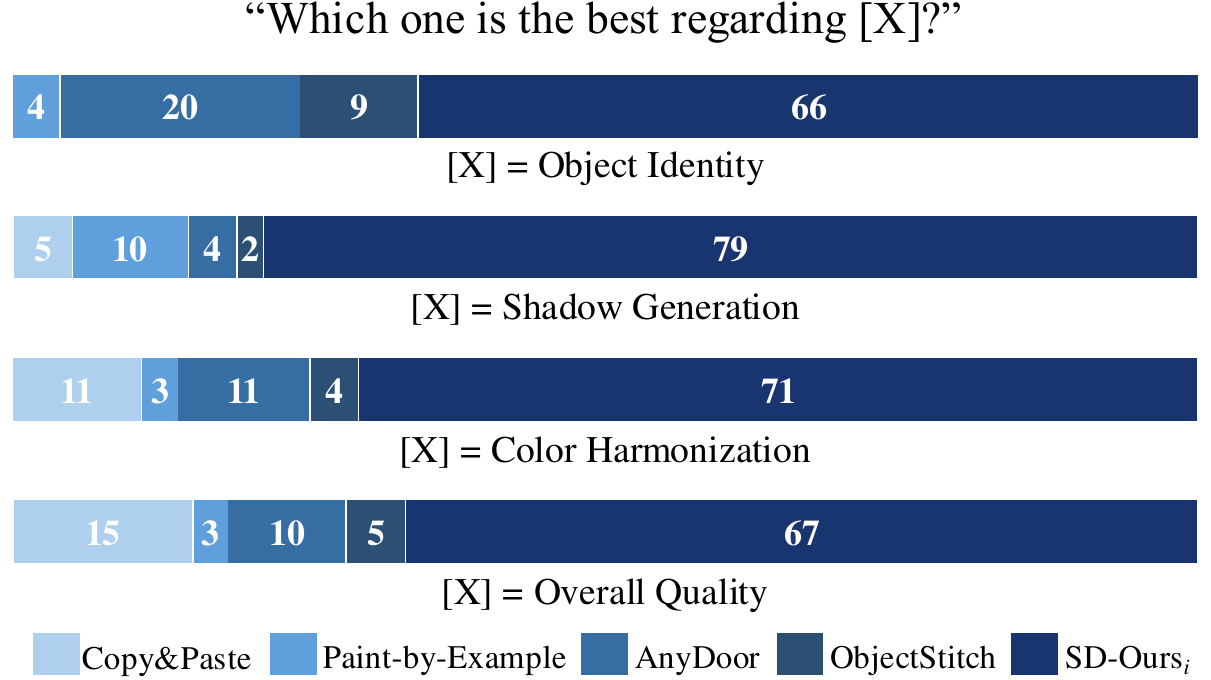}
    \caption{\textbf{Object insertion - user study.} Participants evaluated different methods on four criteria: object identity preservation, shadow generation (object-to-scene effects), color harmonization (scene-to-object effects), and overall quality based on participants' preference. Copy \& Paste method is excluded for the object identity preservation test. Numbers are reported in \%.}
    \label{fig:insertion_userstudy}
\end{figure}

For object insertion task, we compare our model (SD-Ours$_i$) with Copy \& Paste, Paint-by-Example \cite{yang2023paint}, AnyDoor \cite{chen2024anydoor}, and ObjectStitch \cite{song2023objectstitch}. 
As shown in \Cref{fig:insertion}, SD-Ours$_i$ consistently demonstrates effective integration of objects into diverse scenes, showing strengths in identity preservation, shadow generation, and color harmonization.

The Copy \& Paste method, while straightforward, encounters challenges with blending the object naturally into the scene and lacks the capability to generate object-to-scene effects, such as realistic shadows. 
Paint-by-Example offers an enhanced level of generating natural images, however, struggles with maintaining object identity, which can be a critical limitation in object-centric image editing.

AnyDoor and ObjectStitch provide more coherent results overall.
Nevertheless, they sometimes encounter difficulties in fully preserving the object’s identity, adapting its colors seamlessly to the scene and and generating natural shadows. 
In contrast, SD-Ours$_i$ achieves high visual consistency, producing realistic shadows, and context-aware colors that align naturally with the lighting conditions in the target scenes, all without major model modifications.

User study results from 62 participants (\Cref{fig:insertion_userstudy}) further support our claim. 
SD-Ours$_i$ received the highest ratings across all four evaluation criteria: object identity preservation, shadow generation (object-to-scene effects), color harmonization (scene-to-object effects), and overall quality. It achieved preference scores of 66\% for object identity, 79\% for shadow/reflection generation, 71\% for color harmonization, and 67\% for overall quality, significantly outperforming other methods. 
These results highlight our dataset as a valid training resource for achieving realistic and contextually integrated object insertions.
\section{Conclusion}
\label{sec:conclusion}

We introduce ORIDa, the first large-scale, real-captured public dataset for object compositing, with over 30,000 images featuring 200 unique objects. 
ORIDa is both extensive and carefully curated, with high-quality, richly annotated images that have undergone thorough data filtering. 
A unique feature of ORIDa is its capture of both object-to-scene and scene-to-object effects, providing diverse object placements in real-world scenes.

Our analysis suggests that ORIDa can support advancements in object compositing by providing a valuable resource for developing more realistic, context-aware image editing techniques. 
Limitations, future research directions, broader impact, and details of our dataset and experiments are provided in the supplementary materials.
We hope ORIDa’s accessibility will inspire further exploration and constructive discussions on compositional image editing within the research community.

\clearpage
\renewcommand{\thesection}{\Alph{section}}
\renewcommand{\thefigure}{\Alph{figure}}
\renewcommand{\thetable}{\Alph{table}}
\setcounter{section}{0}
\setcounter{table}{0}
\setcounter{figure}{0}

\maketitlesupplementary

\section{Additional Data Samples}
\label{supp_sec:data_samples}


\subsection{Dataset Samples}
\begin{figure*}[htbp]
    \centering
    \includegraphics[width=\textwidth]{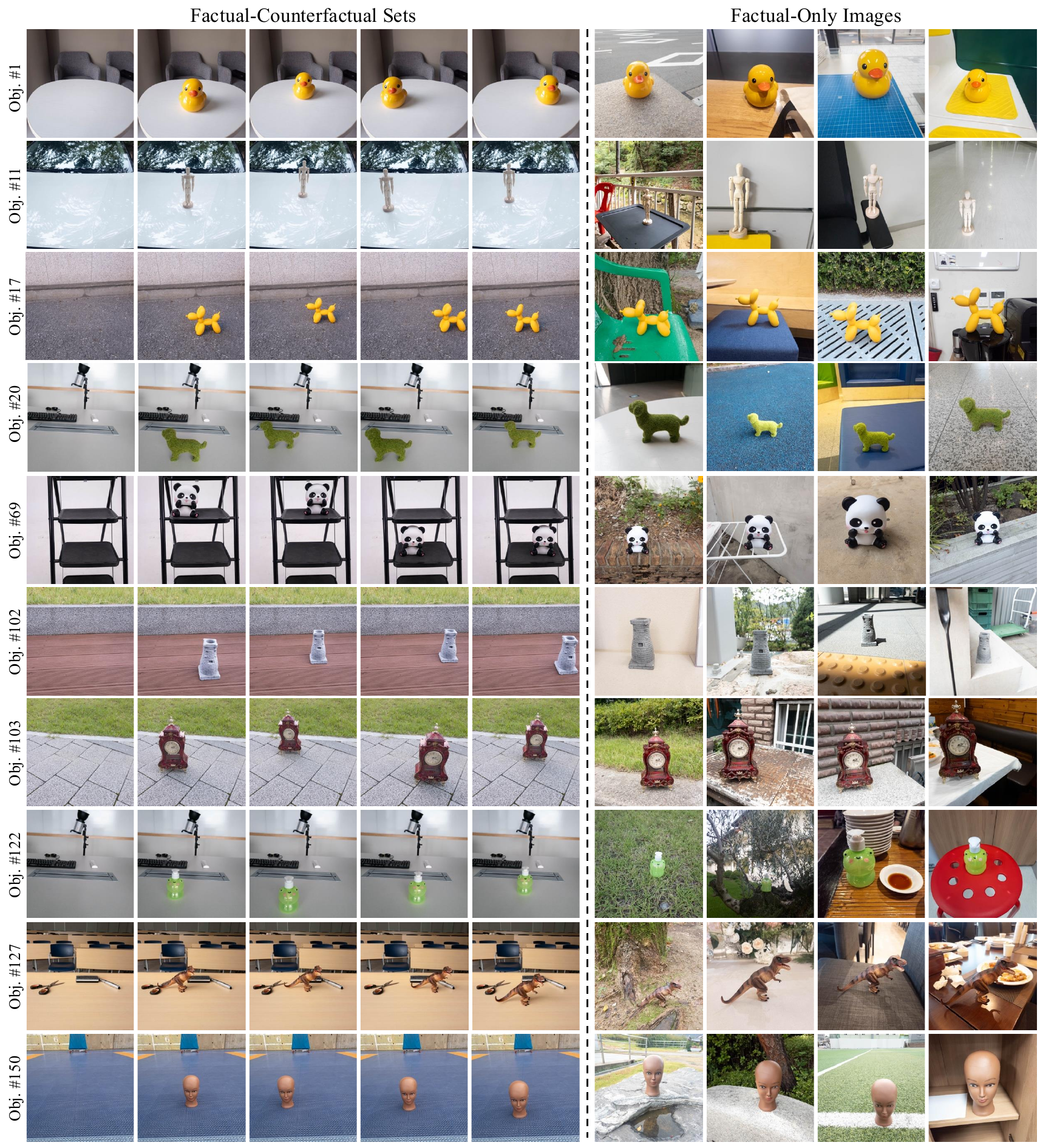}
    \vspace{-0.6em}
    \caption{Additional examples of Factual-Counterfactual (F-CF) sets and Factual-Only (F-Only) images.}
    \label{supp_fig:data_samples}
\end{figure*}

\begin{figure*}[htbp]
    \centering
    \includegraphics[width=\textwidth]{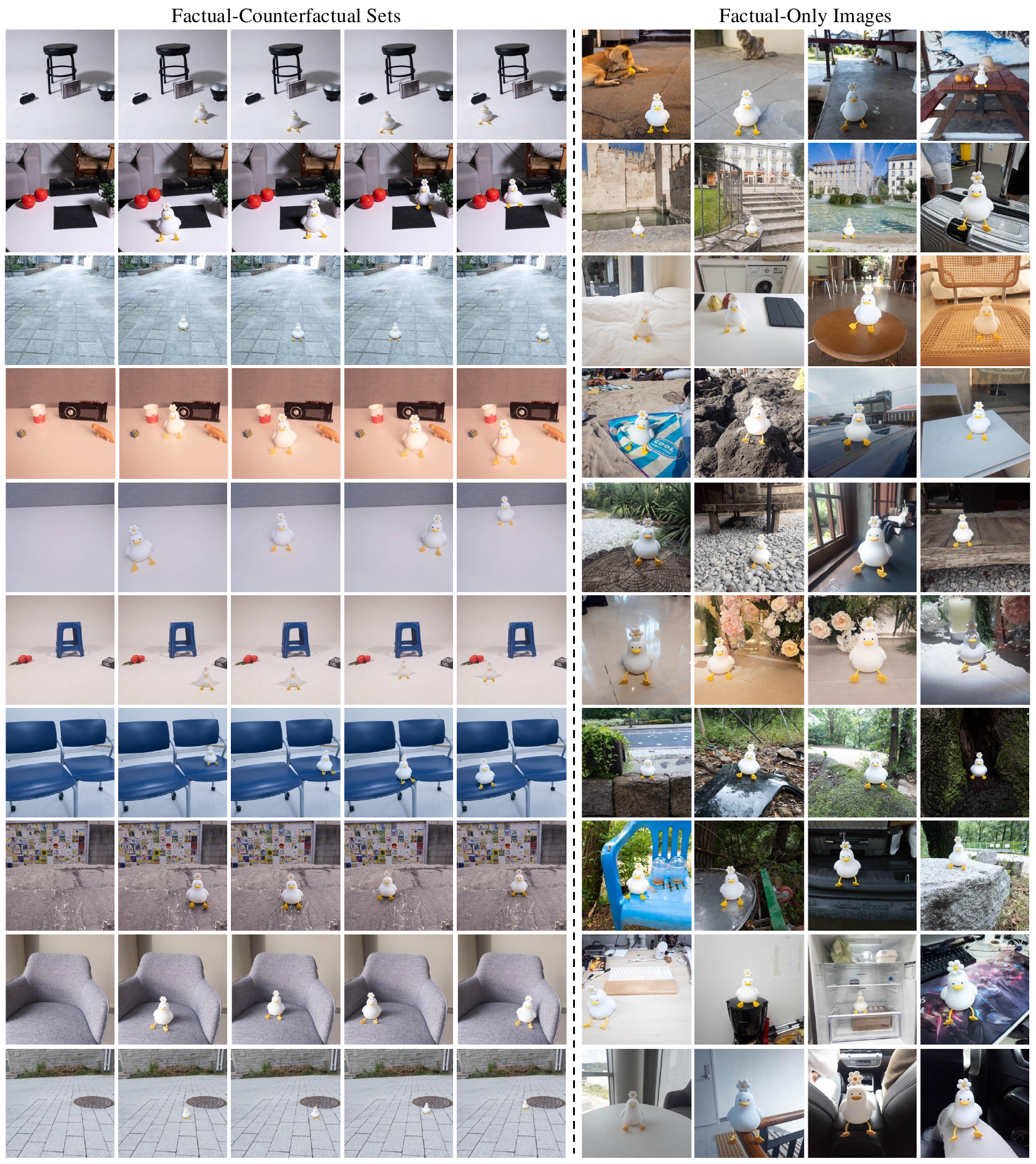}
    \vspace{-0.6em}
    \caption{Additional examples of Factual-Counterfactual (F-CF) sets and Factual-Only (F-Only) images for object \#2.}
    \label{supp_fig:data_samples2}
\end{figure*}

\Cref{supp_fig:data_samples} and \Cref{supp_fig:data_samples2} display examples of Factual-Counterfactual (F-CF) sets and Factual-Only (F-Only) images.
\Cref{supp_fig:data_samples} includes multiple objects, where F-CF sets (left) contain multiple object positions and their corresponding background-only images, while F-Only images (right) show object variations in different scenes.
\Cref{supp_fig:data_samples2}, on the other hand, focuses on a single object (object \#2), illustrating its placement across various contexts, emphasizing the dataset’s ability to capture real-world diversity for individual objects.

\subsection{Object Categories}
\begin{figure*}[htbp]
    \centering
    \includegraphics[width=0.98\textwidth]{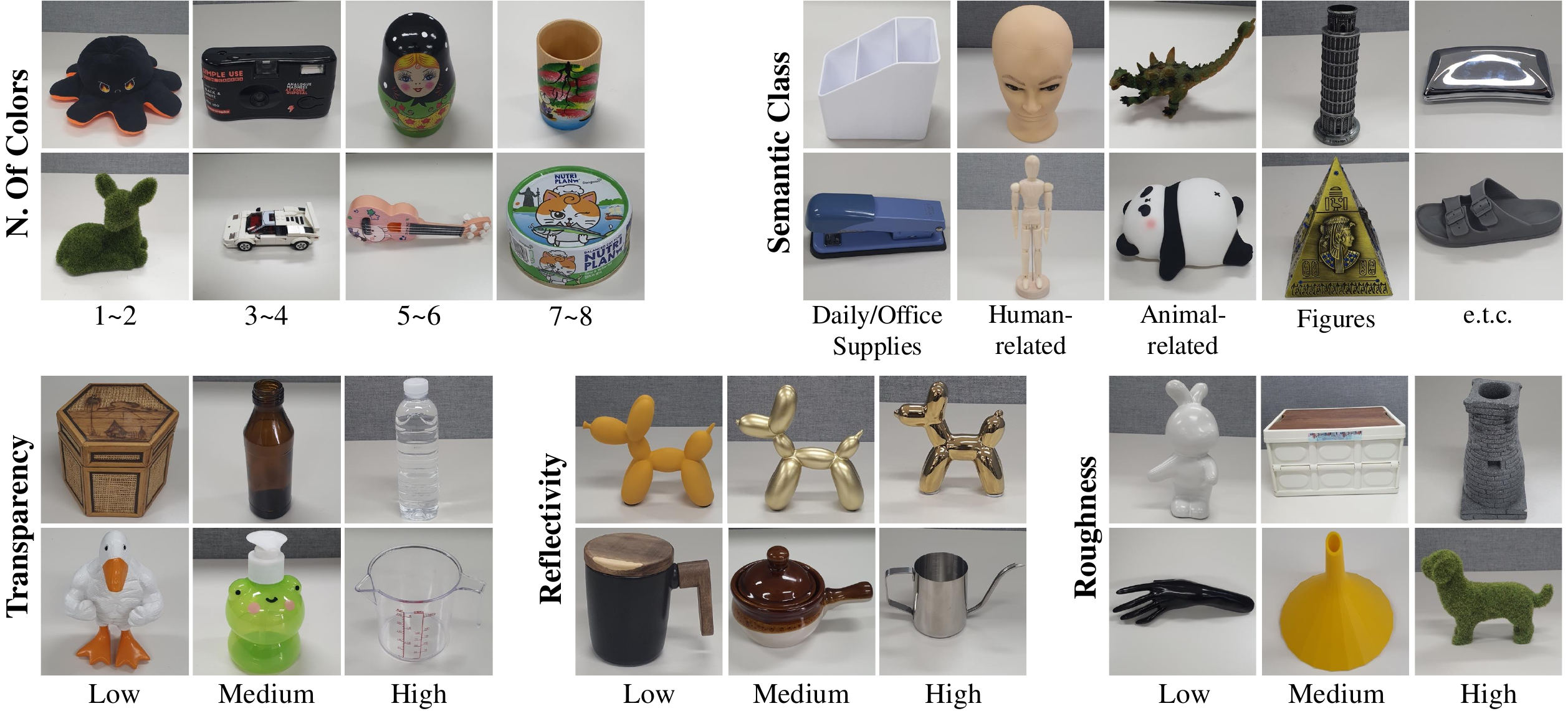}
    \vspace{-0.2em}
    \caption{
        Examples of objects categorized by number of colors, semantic class, transparency, reflectivity, and surface roughness.
    }
    \label{supp_fig:category_samples}
\end{figure*}

\Cref{supp_fig:category_samples} illustrates the diversity of ORIDa through objects categorized by attributes such as the number of colors, semantic class, transparency, reflectivity, and roughness. 
Objects are grouped by color complexity (1–2 to 7–8 colors) and semantic classes, including daily/office supplies, human-related items, animal-related objects, figures, and miscellaneous categories. 
Transparency, reflectivity, and roughness levels further represent diverse material properties and surface textures, capturing variations in light interaction and texture.
These categories reflect the thoughtful curation of ORIDa, enabling support for a broad range of image editing tasks.

\subsection{ISP-augmented data}
\Cref{supp_fig:data_samples} also demonstrates images processed through five distinct ISP settings using Adobe Lightroom Classic. 
These settings include: (1) as-shot (default settings), (2) higher temperature, (3) lower temperature, (4) higher vibrance, and (5) lower vibrance. 
These augmentations expand the dataset’s variety, enabling better generalization for training models under varying lighting and color conditions.

\section{Experimental Details}
\label{supp_sec:experimental_details}

\subsection{Train Schedule}
We fine-tune our models starting from the pre-trained SD-Inpaint model \cite{rombach2022high,huggingface2024diffusersinpainting} using the Adam optimizer \cite{kingma2014adam} with a learning rate of 5e-5 and a cosine scheduler \cite{loshchilov2016sgdr}.
The batch size is set to 64 for both object removal and insertion tasks.

For object removal, the model is fine-tuned for 5,000 steps, resulting in 320,000 training samples, significantly fewer than ObjectDrop’s 6.4 million samples, generated over 50,000 steps with a batch size of 128.
For object insertion, we train the model for 500,000 steps, resulting in 32 million training samples.
This is still fewer than ObjectDrop’s 56.3 million samples, generated through a two-stage process: 100,000 iterations with a batch size of 512 on synthetic data, followed by 40,000 iterations with a batch size of 128 on real-captured data \cite{winter2024objectdrop}.
It is also fewer than Paint-by-Example’s 76 million samples, produced over 40 epochs using a synthetic dataset of 1.9 million images \cite{yang2023paint}.
Training the object insertion model takes about 150 hours using 4 NVIDIA A100-PCIE (40GB) GPUs.

\subsection{Model Inputs}  
Our framework is built upon the pre-trained SD-Inpaint model \cite{rombach2022high,huggingface2024diffusersinpainting}, where the U-Net processes a 9-channel input: four channels for the input latent, four for the condition latent, and one for the object mask.  

\noindent \textbf{Object Removal.}  
For training, the input latent is a perturbed version of the original image's latent representation
The condition latent is created by masking the input latent using the object mask.  
During inference, the model inputs for object removal remain identical to the training setup.  

\noindent \textbf{Object Insertion.}  
For training, the input latent is the perturbed latent of the ground truth (real-captured) object-included image. 
The condition latent is the latent of a Copy \& Paste image, which is generated by masking and pasting the source object into the target image.  
During inference, as the ground truth object-included image is unavailable, we use the Copy \& Paste image as the input latent. 
All other settings remain consistent with the training configuration.  

\subsection{Diffusion Model}  
We primarily follow the pipeline of SD-Inpaint \cite{huggingface2024diffusersinpainting}, making minimal modifications only during the inference stage.  
To better preserve the source object’s identity in the object insertion task, we employ \textit{skip residual} connections inspired by DemoFusion \cite{du2024demofusion}. 
This method combines the noised latent reference \( z'_t \sim q(z_t | z_0) \), derived from the original input image’s latent representation \( z_0 \), with the current denoised latent \( z_t \sim p_\theta(z_t | z_{t+1}) \). 
The contributions of \( z'_t \) and \( z_t \) are dynamically weighted using a cosine scheduling mechanism, which adjusts the balance between the two throughout the denoising process. 
For more details on this weighting approach, refer to the original paper \cite{du2024demofusion}.  

By leveraging information from \( z_0 \), the model retains the source object’s identity while seamlessly blending it into the target scene. 
Apart from this inference-stage adjustment using skip residuals, the underlying SD-Inpaint model remains unchanged.

\section{Additional Experiments}
\label{supp_sec:results}

\subsection{Object Removal}
\begin{figure*}[htbp]
    \centering
    \includegraphics[width=0.92\textwidth]{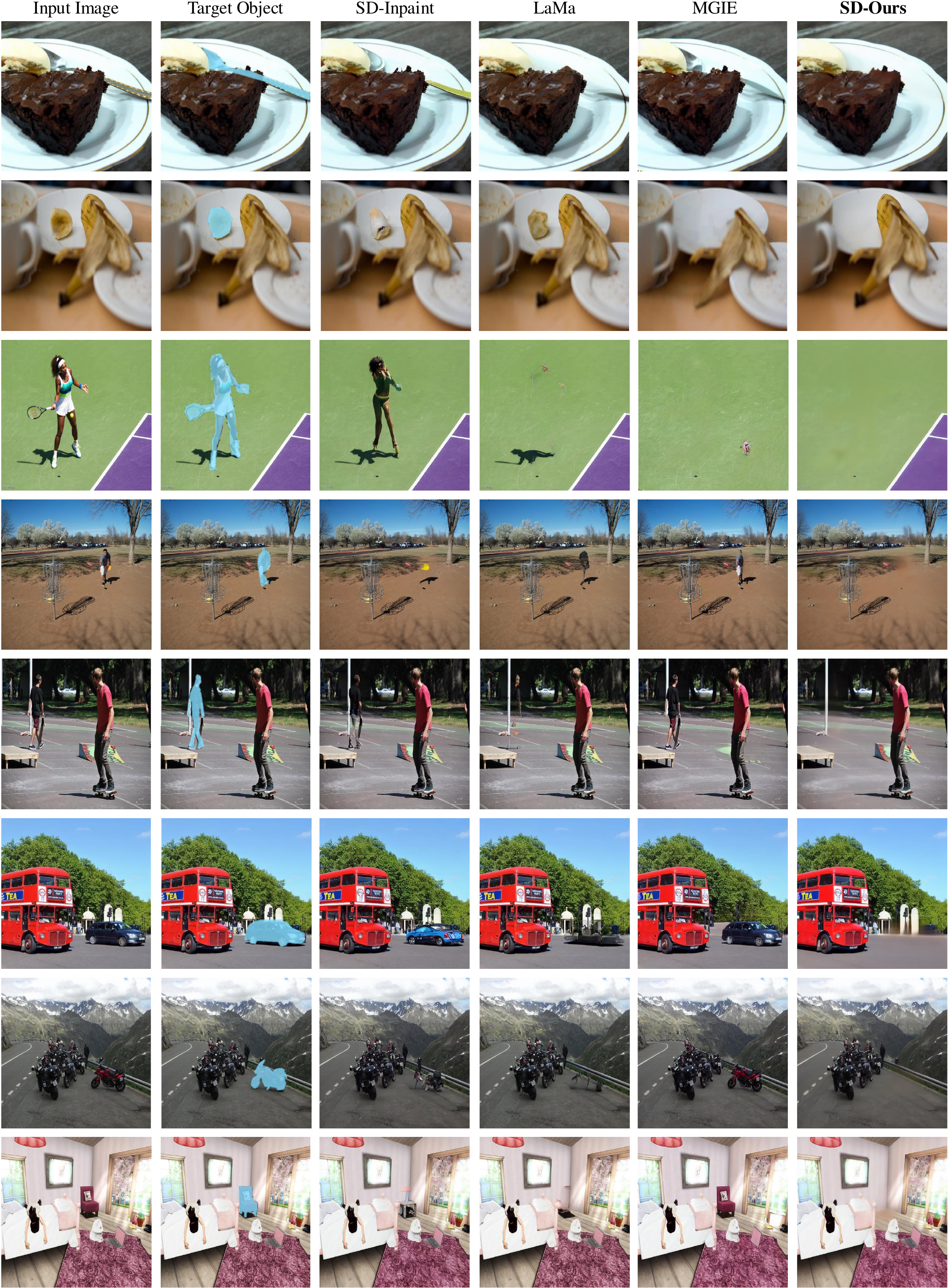}
    \vspace{-0.6em}
    \caption{Additional results of object removal with various models.}
    \label{supp_fig:removal}
\end{figure*}

\Cref{supp_fig:removal} presents further object removal results using multiple models, including SD-Inpaint \cite{rombach2022high,huggingface2024diffusersinpainting}, LaMa \cite{suvorov2022resolution}, MGIE \cite{fu2023guiding}, and SD-Ours$_{r}$. 
The examples highlight how different methods handle challenges such as background reconstruction, shadow removal, and artifact elimination.

\subsection{Object Insertion}
\begin{figure*}[htbp]
    \centering
    \includegraphics[width=\textwidth]{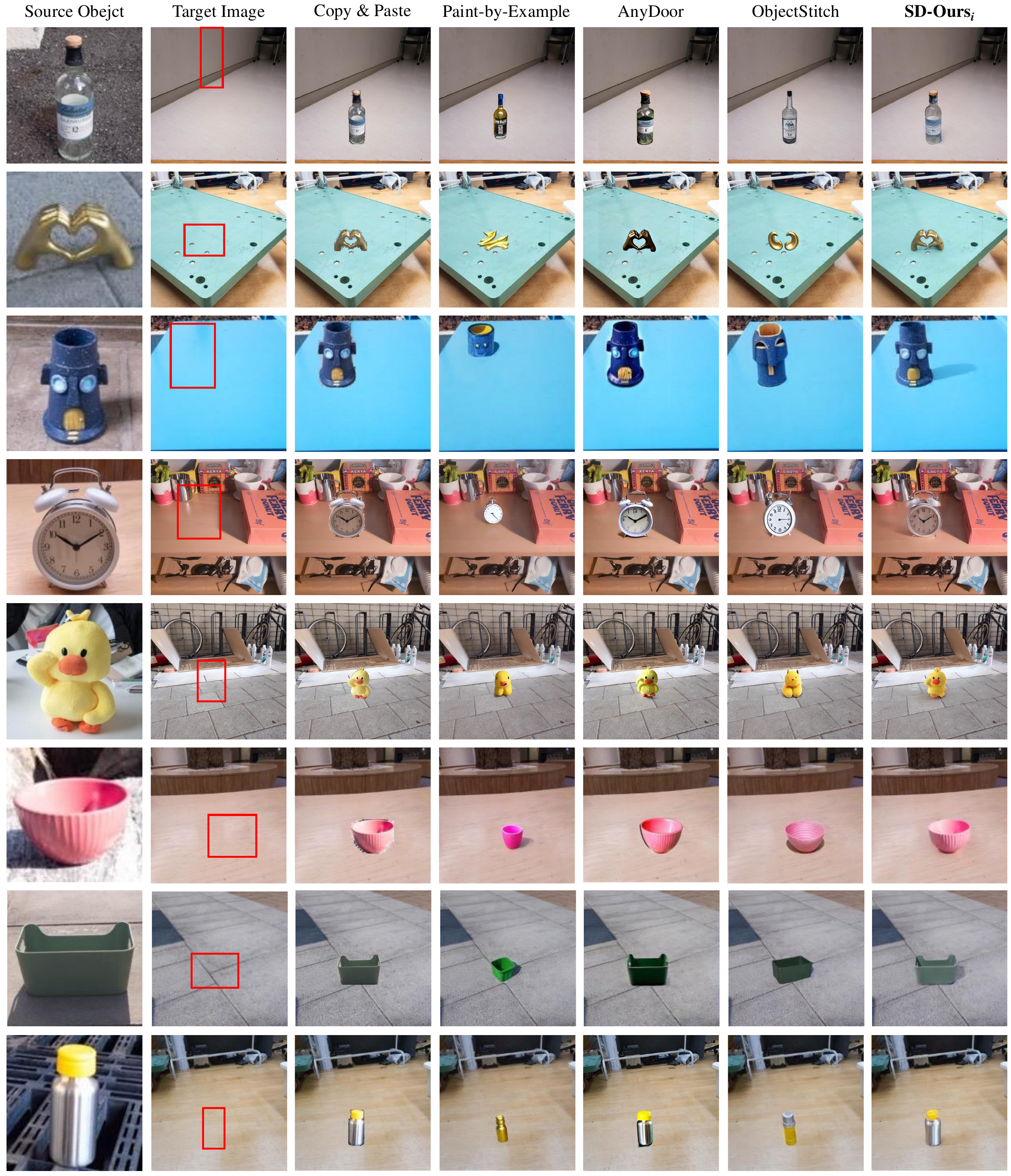}
    \vspace{-0.6em}
    \caption{Additional results of object removal generated by various models.}
    \label{supp_fig:insertion}
\end{figure*}

\begin{figure*}[htbp]
    \centering
    \includegraphics[width=\textwidth]{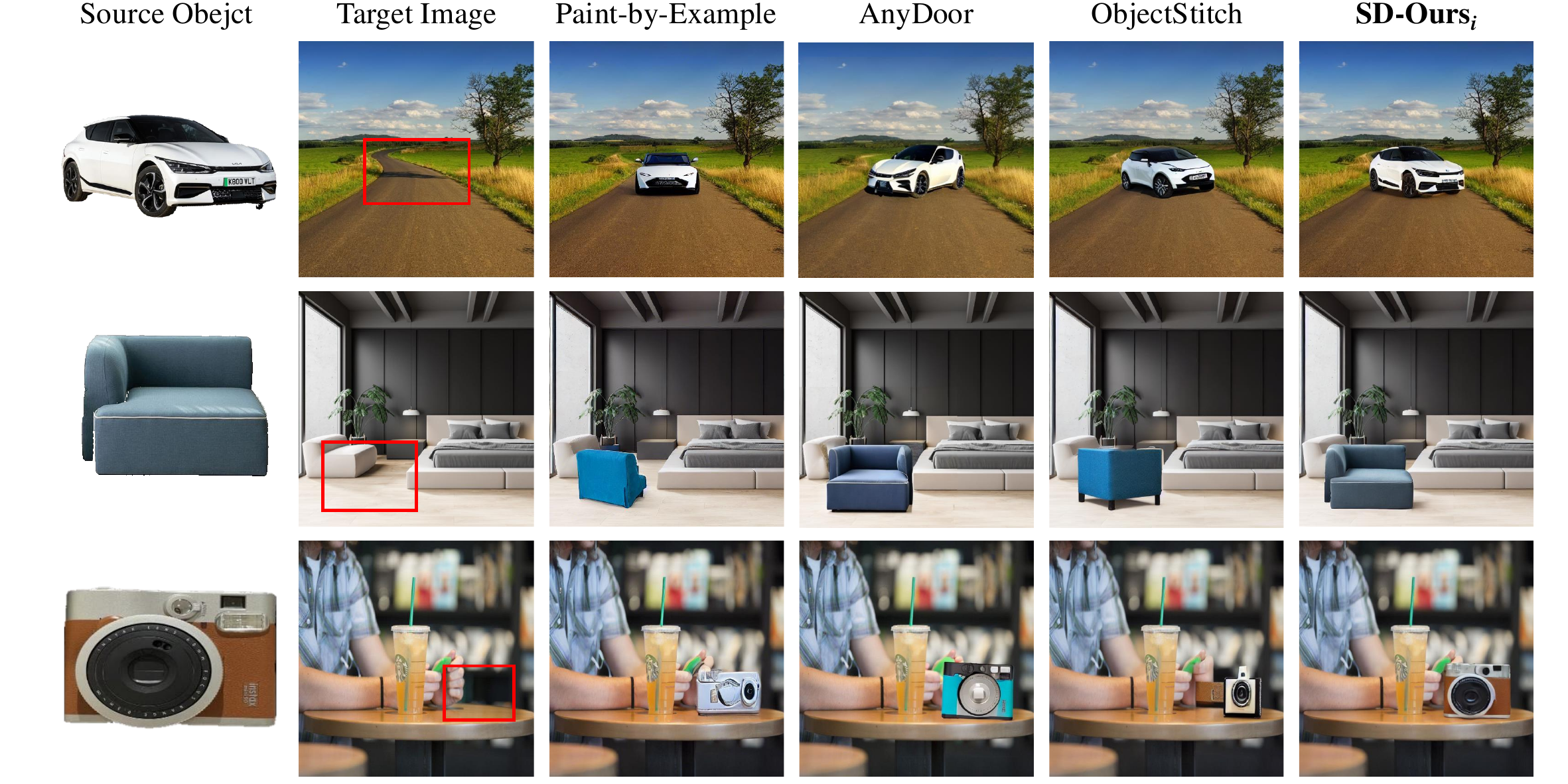}
    \vspace{-0.6em}
    \caption{Additional results for object insertion with objects and scenes from internet and an external dataset.}
    \label{supp_fig:insertion2}
\end{figure*}

Additional qualitative results for object insertion are shown in \Cref{supp_fig:insertion} with ORIDa test set and \Cref{supp_fig:insertion2} with in-the-wild data from internet and an external dataset, MureCom \cite{chen2024mureobjectstitch}.
The results demonstrate the effectiveness of different models -- Copy \& Paste, Paint-by-Example \cite{yang2023paint}, AnyDoor \cite{chen2024anydoor}, ObjectStitch \cite{song2023objectstitch}, and SD-Ours$_{i}$ -- in maintaining object identity, harmonizing colors, and generating shadows and reflections for seamless integration.

\subsection{Ablation Study on Data Scale}
\begin{figure*}[htbp]
    \centering
    \includegraphics[width=\textwidth]{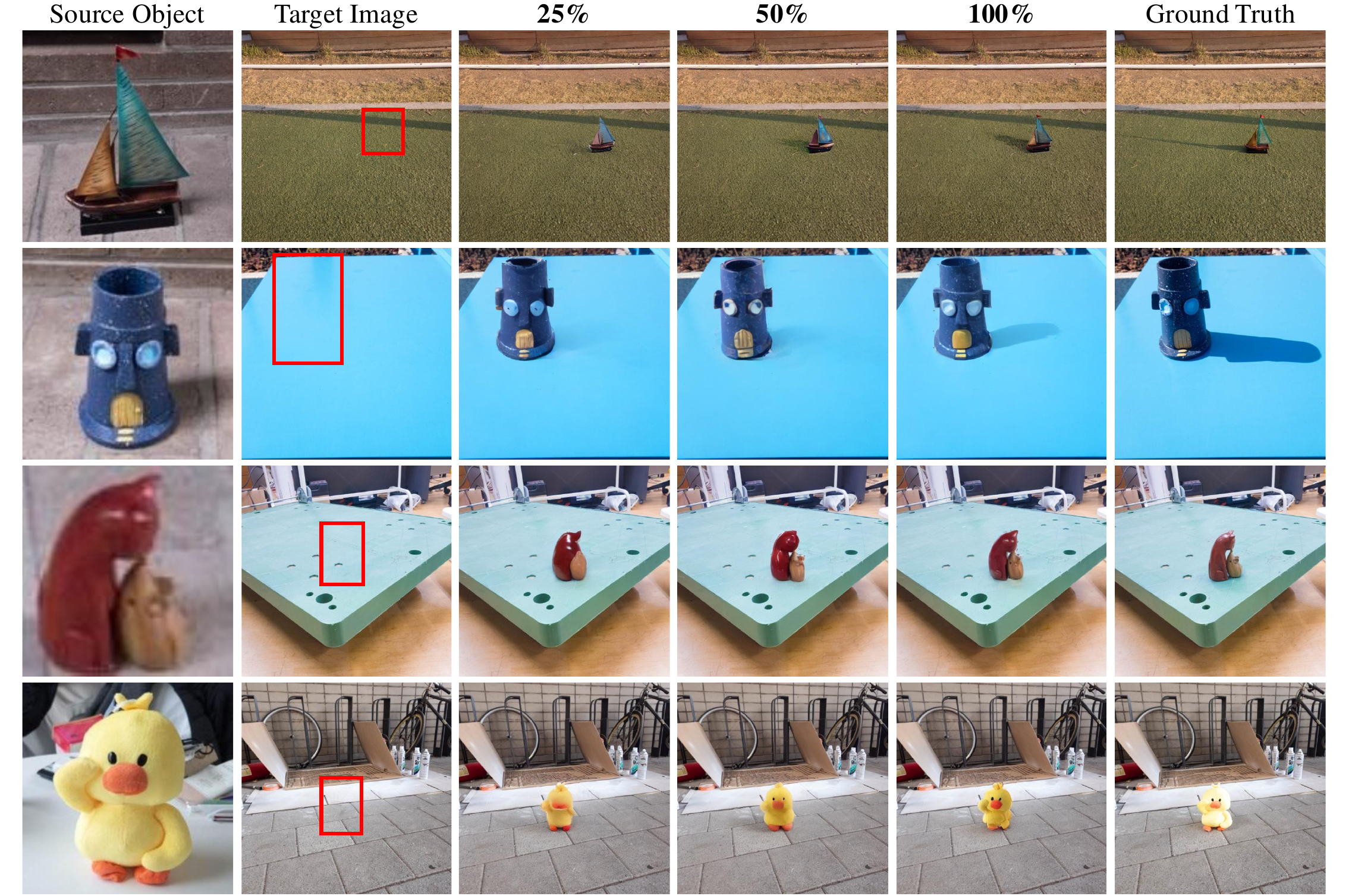}
    \vspace{-0.6em}
    \caption{Ablation study results on the effect of training dataset size for the object insertion task.}
    \label{supp_fig:ablation_insertion}
\end{figure*}

We performed an ablation study to evaluate the impact of dataset scale (25\%, 50\%, 100\%) on object insertion performance, focusing on shadow generation and source object preservation (\Cref{supp_fig:ablation_insertion}). 
Models trained on 25\% of the dataset struggled with context-aware shadow generation and exhibited artifacts in object appearance. 
Increasing to 50\% improved performance, however, some inconsistencies still remained. 
Training on the full dataset (100\%) yielded the best results, with accurate shadows and faithful preservation of object identity and appearance. 
This demonstrates the importance of dataset scale for achieving high-quality, context-aware object insertion.
\section{Limitations and Future Works}
\label{supp_sec:limitations}


\noindent \textbf{Excluded object types.}
Our dataset excludes certain categories of objects to maintain consistency and feasibility during data collection. 
First, human subjects are excluded due to complexities related to appearance variability and ethical considerations. 
Additionally, we have excluded objects characterized by significant temporal variability (e.g., living organisms, perishable food items, and deformable or flexible materials), as well as large-scale objects impractical for repeated captures. 
Addressing these exclusions in future dataset versions could significantly expand the range of applicable research and practical scenarios.

\noindent \textbf{Dataset scale-up.} 
We introduced a dataset for object-centric image composition at an unprecedented scale, containing 200 unique objects across 30,000 images. Despite this significant advancement, it remains insufficient to fully represent the vast diversity and complexity of real-world visual scenarios. We envision that both our dataset and the methodologies developed to capture it will serve as foundational resources for future datasets. An important direction is to simplify and streamline our data collection process, enabling scalable crowd-sourced dataset acquisition.

\noindent \textbf{Limited 3D information and pose variability.}
Our dataset does not include explicit 3D information, such as multi-view captures or ground truth depth maps. Additionally, the dataset intentionally restricts variation in object poses to maintain consistency and highlight object placement across scenes, resulting in limited diversity in pose dynamics. Future datasets might address this limitation by incorporating more comprehensive pose variations and additional 3D-related annotations.


\noindent \textbf{Limited novelty in model development.} 
To emphasize the value of our high-quality real-world dataset, which can directly serve as training samples for advanced diffusion models, we primarily employed vanilla models \cite{rombach2022high,huggingface2024diffusersinpainting} with minimal modifications. Consequently, our research did not explore potential performance improvements achievable through advanced architectural innovations or customized model enhancements. Future work may benefit from integrating novel model architectures specifically tailored for object-centric image composition tasks.




\section{Broader Impact}
\label{sec:broader_impact}

ORIDa advances realistic image compositing, supporting augmented/virtual reality and AI-driven content production. 
However, its realism raises concerns about misuse, such as deepfakes or deceptive media. 
We encourage responsible use and adherence to ethical guidelines to balance innovation with societal safeguards.

\clearpage

\begin{figure*}[htbp]
    \centering
    \includegraphics[width=\linewidth]{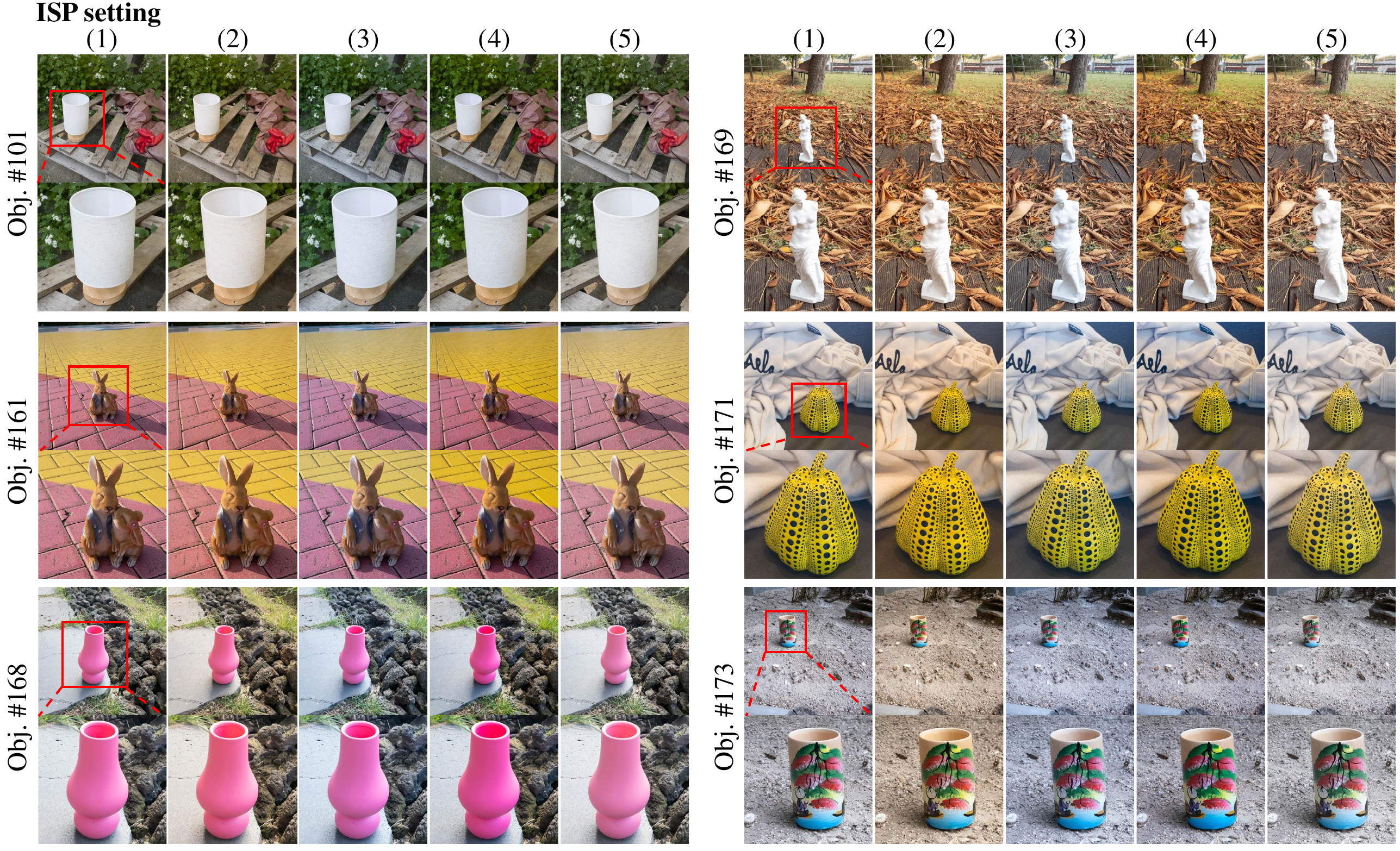}
    \vspace{-1em}
    \caption{
        Example images data for five different ISP settings:
        (1) as-shot, (2) higher temperature, (3) lower temperature, (4) higher vibrance, and (5) lower vibrance. 
    }
    \label{supp_fig:isp_samples}
\end{figure*}

\clearpage

{
    \small
    \bibliographystyle{ieeenat_fullname}
    \bibliography{main}
}

\end{document}